\documentclass[10pt,twocolumn,letterpaper]{article}

\usepackage[pagenumbers]{cvpr} 

\usepackage{times}
\usepackage{microtype}
\usepackage{epsfig}
\usepackage{float}
\usepackage{placeins}
\usepackage{stfloats}
\usepackage{enumitem}
\usepackage{tabularx}
\usepackage{xstring}
\usepackage{multirow}
\usepackage{xspace}
\usepackage{url}
\usepackage{subcaption}
\usepackage{xcolor}
\definecolor{DarkGreen}{rgb}{0.43, 0.68, 0.28}

\def\ie{\emph{i.e}\onedot} 
\newcommand{\vidframe}{v}
\newcommand{\flow}{f}

\newcommand{\lossmain}{\mathcal{L}_{\text{main}}}
\newcommand{\lossbce}{\mathcal{L}_{\text{bce}}}
\newcommand{\lossssl}{\mathcal{L}_{\text{ssl}}}
\newcommand{\lossdepth}{\mathcal{L}_{\text{depth}}}

\newcommand{\sslweight}{\lambda}
\newcommand{\encoder}{\mathcal{E}}
\newcommand{\modulation}{\mathcal{M}}

\newcommand{\enimage}{\mathcal{E}_{v}}
\newcommand{\enflow}{\mathcal{E}_{f}}

\newcommand{\demain}{\mathcal{D}_{main}}
\newcommand{\demask}{\mathcal{D}_{m}}
\newcommand{\dessl}{\mathcal{D}_{ssl}}
\newcommand{\dedepth}{\mathcal{D}_{d}}

\newcommand{\inframe}{v}
\newcommand{\inflow}{f}

\newif\ifreview 
\newif\ifarxiv 
\newif\ifcamera 
\newif\ifrebuttal 

\def\paperID{8392} 
\def\confName{CVPR}
\def\confYear{2024}

\definecolor{cvprblue}{rgb}{0.21,0.49,0.74}
\usepackage[pagebackref,breaklinks,colorlinks,citecolor=cvprblue]{hyperref}

\def\paperID{8392} 
\def\confName{CVPR}
\def\confYear{2024}

\title{Depth-aware Test-Time Training for Zero-shot Video Object Segmentation}

\author{
    Weihuang Liu\textsuperscript{\rm 1} ~
    Xi Shen\textsuperscript{\rm 2} ~
    Haolun Li\textsuperscript{\rm 1} ~
    Xiuli Bi\textsuperscript{\rm 3} ~
    Bo Liu\textsuperscript{\rm 3} ~
    Chi-Man Pun\textsuperscript{\rm 1,}\thanks{Corresponding Author} ~
    Xiaodong Cun\textsuperscript{\rm 4,}\footnotemark[1] 
    \\
    \textsuperscript{\rm 1}~University of Macau \qquad 
    \textsuperscript{\rm 2}~Intellindust \qquad 
    \\
    \textsuperscript{\rm 3}~Chongqing University of Posts and Telecommunications \qquad 
    \textsuperscript{\rm 4}~Tencent AI Lab 
}

\begin{document}
\maketitle
\begin{abstract}

Zero-shot Video Object Segmentation~(ZSVOS) aims at segmenting the primary moving object without any human annotations. 
Mainstream solutions mainly focus on learning a single model on large-scale video datasets, which struggle to generalize to unseen videos. In this work, we introduce a test-time training~(TTT) strategy to address the problem. Our key insight is to enforce the model to predict consistent depth during the TTT process. In detail, we first train a single network to perform both segmentation and depth prediction tasks. This can be effectively learned with our specifically designed depth modulation layer. Then, for the TTT process, the model is updated by predicting consistent depth maps for the same frame under different data augmentations. In addition, we explore different TTT weight updating strategies. Our empirical results suggest that the momentum-based weight initialization and looping-based training scheme lead to more stable improvements. Experiments show that the proposed method achieves clear improvements on ZSVOS. Our proposed video TTT strategy provides significant superiority over state-of-the-art TTT methods. 
Our code is available at:~\url{https://nifangbaage.github.io/DATTT/}.

\end{abstract}
\section{Introduction}
\label{sec:intro}

\begin{figure}[t]
\centering
\includegraphics[width=\linewidth]{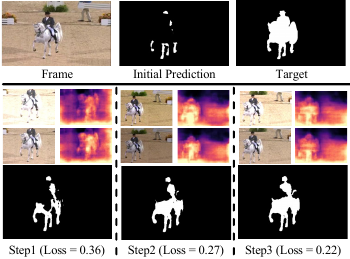}
\caption{\textbf{Key idea of our  Depth-aware Test-Time Training.} During the test-time training, the model is required to predict consistent depth maps for the same video frame under different data augmentation (2nd row). The model is progressively updated and provides more precise mask prediction (3rd row). 
}
\label{fig:intro}
\end{figure}

Zero-shot video object segmentation~(ZSVOS) is a fundamental task in computer vision, aiming to accurately segment the primary moving object within a video. The term ``zero-shot" means no human guidance is provided during the inference, which is different from one-shot video object segmentation~(OSVOS), where the annotation of the first frame is provided. The problem is of great importance due to its wide applications in video understanding~\cite{liu2021tam}, video surveillance~\cite{hou2019vrstc}, video editing~\cite{zhang2023sadtalker,fatezero}, \etc.

However, this task remains challenging, as the model is required to learn category-agnostic features to detect moving objects. 
Despite being trained on the most widely used public dataset~($\sim$3.5K videos), the deployed models often fail when confronted with real-world scenarios. The above issue might be attributed to the lack of large-scale training sets. However, collecting and annotating large-scale video datasets is costly. As an alternative solution, test-time training~(TTT) has emerged as a promising approach, which serves as the main focus of this work. TTT consists of conducting training on each test video. Therefore, the model is expected to automatically adapt to new scenarios.

Leading studies~\cite{sun2020test,liu2021ttt++,gandelsman2022test} on TTT are mainly focused on image recognition and have showcased that a well-designed self-supervised objective function enables the model to adapt to new distributions, thereby achieving an improved fit for individual test samples. These approaches are then extended to videos for OSVOS~\cite{azimi2022self} and video semantic segmentation~\cite{volpi2022road,wang2023test}, and demonstrated to be effective on unusual weather scenarios~(snow, fog, rain, \etc). Different from these works, our key idea in this work (see Figure~\ref{fig:intro}) is to leverage depth information on video TTT. Our motivation comes from the fact that the primary object should be close to the camera and thus has a relatively small depth, \ie the depth map should contain informative signals to segment primary moving objects.

In this study, we propose a novel framework named Depth-aware Test-Time Training (DATTT) for ZSVOS. Similar to other video TTT frameworks~\cite{azimi2022self,volpi2022road,wang2023test}, our DATTT is a two-stage training framework. In the first stage, we learn a single model that jointly predicts the mask of the moving object as well as the depth map of the entire image. 
To achieve this, the model is built upon the commonly used two-stream ZSVOS model by incorporating a depth decoder that utilizes image features to predict the depth map. 
Therefore, the model has a shared image encoder, a flow encoder, and different decoder heads for each task. 
We also find that the interaction between these two tasks through a depth modulation layer enables better performance when performing TTT. 
In the second TTT stage, given an input video, the model is required to predict consistent depth maps under different data augmentations for each frame. Through optimizing the consistency loss, the image encoder part of the model is updated, leading to adaptive predictions on the mask head. Note that the depth supervision in the first stage comes from monocular depth predictors~\cite{godard2019digging,zhang2023lite,bhat2023zoedepth}, thus it is free but assumed to be noisy.  

Experimentally, we evaluate our DATTT on five widely-used ZSVOS datasets: DAVIS-16~\cite{perazzi2016benchmark}, FBMS~\cite{ochs2013segmentation},
Long-Videos~\cite{liang2020video}, MCL~\cite{kim2015spatiotemporal}, and SegTrackV2~\cite{li2013video}. Our empirical results suggest that both the first stage training and the TTT training benefit from the additional depth information introduced in the network. The improvement is clear and consistent across different depth predictors and the proposed depth modulation layer also provides important performance gain for TTT. We also explore different TTT strategies and ultimately find that the momentum-based weight initialization and looping-based training scheme lead to more consistent improvement. In terms of fair comparison to competitive TTT approaches~\cite{schneider2020improving,wang2020tent,sun2020test,gandelsman2022test}, our DATTT enables more stable improvement for TTT and manages to achieve significantly better performance. Although  very different from state-of-the-art ZSVOS approaches, our DATTT still provides competitive performance, demonstrating the effectiveness of conducting TTT during inference.

To summarize, our main contributions are as follows:
\begin{itemize}
\item We introduce the Depth-aware Test-Time Training (DATTT) for zero-shot video object segmentation (ZSVOS). To the best of our knowledge, for the first time, demonstrating that performing TTT with consistent depth constraint brings significant improvement.

\item We propose a depth modulation layer which enables interaction between the depth prediction head and mask prediction head and has shown to be effective for TTT process.

\item Our DATTT achieves competitive performances compared to state-of-the-art approaches on ZSVOS, demonstrating the effectiveness of performing TTT during inference. 

\end{itemize}
\section{Related Work}
\label{sec:related}

\noindent\textbf{Zero-shot Video Object Segmentation.}
Zero-shot video object segmentation (ZSVOS) is a task aimed at segmenting the primary moving objects without requiring any annotations during inference. Traditionally, heuristic algorithms, including background subtraction~\cite{zivkovic2006efficient,culibrk2007neural}, object proposal~\cite{lee2011key,zhang2013video}, and point trajectories~\cite{brox2010object,ochs2012higher}, have been commonly employed to address ZSVOS. 
With the rapid advancements in deep learning, neural networks have emerged as the most popular technique for ZSVOS~\cite{wang2019zero,tokmakov2019learning,zhuo2019unsupervised}.
To exploit the temporal information in the video sequence, early works~\cite{tokmakov2017learning,wang2019learning,ventura2019rvos} have developed recurrent neural network~(RNN)-based models to leverage the correlations between successive frames. 
More recent research~\cite{yang2021learning,zhang2021deep,pei2022hierarchical,pei2023hierarchical,cho2023treating} has focused on incorporating motion information, resulting in significant performance improvements. Off-the-shelf optical flow estimation methods~\cite{sun2018pwc,teed2020raft} are utilized to extract motion cues, which are then combined with appearance information in a two-stream model.
For example, Yang \etal~\cite{yang2021learning} propose an attentive multi-modality collaboration network that integrates appearance and motion information using a co-attention mechanism. This multi-modality feature fusion suppresses misleading information and emphasizes the relevant foreground features. Since the optical flow which represents motion information for all pixels often fails to align well with the primary objects,  Pei \etal~\cite{pei2022hierarchical} introduce a hierarchical feature alignment network that aligns appearance and motion features hierarchically using distinct modules.

\noindent\textbf{Depth-based Object Segmentation.}
The depth map obtained by depth sensors offers valuable geometric insights for scene understanding. Multi-modal features extracted from RGB images and depth maps provide complementary information in both appearance and spatial position.
Depth map has been shown to be beneficial in salient object detection~\cite{chen2018progressively,zhou2021specificity,piao2019depth,sun2021deep,zhao2019contrast,piao2020a2dele, cun2020defocus} since it provides discriminative information in spatial structure. 
Chen \etal~\cite{chen2018progressively} propose a complementarity-aware fusion module to exploit cross-modal information.
Zhao \etal~\cite{zhao2019contrast} propose a contrast-enhanced network to bridge the RGB features and depth features and measure the contrast between salient and non-salient regions.
Liu \etal~\cite{liu2020learning} fuse the multi-modal information via spatial attention.
The integration of depth information as input may hinder the practical application. 
To address this,
Piao \etal~\cite{piao2020a2dele} propose a depth distiller that transfers the depth knowledge to the RGB stream during training and only RGB input is necessary for testing.
To the best of our knowledge, the use of depth maps in video object segmentation remains to be explored.

\noindent\textbf{Test-Time Training.}
Previous works on test-time training~\cite{tonioni2019learning,tonioni2019real,wang2020tent,schneider2020improving,sun2020test,liu2021ttt++,gandelsman2022test} have demonstrated that fine-tuning the pre-trained model for individual instances enables better adaptation to each specific instance. 
Wang \etal~\cite{wang2020tent} minimize test entropy to adapt the normalization layers.
Schneider \etal~\cite{schneider2020improving} replace the trained statics of the normalization layer by estimating from the test sample.
Sun \etal~\cite{sun2020test} develop a Y-shape model containing the backbone, main head, and auxiliary head. During testing, they fine-tune the backbone by predicting the rotation degree using the auxiliary head. 
TTT-MAE \cite{gandelsman2022test} reconstructs the test image using the masked autoencoder to adapt the model to a new test distribution.
Test-time training is also used in some one-shot video object segmentation~(OSVOS) methods~\cite{caelles2017one,voigtlaender2017online,ci2018video}. These kinds of methods further retrain the pre-trained model on a frame with annotation and then test it on the entire video sequence.
Caelles \etal~\cite{caelles2017one} propose the first online training-based OSVOS method. They first finetune the pre-trained model on the video with the first frame annotation, then test the entire video sequence using the new weights.
Since the appearance of the object changes over time, Voigtlaender \etal~\cite{voigtlaender2017online} propose an online adaptation scheme. For each frame, they generate pseudo labels using the estimated masks and a threshold to adapt the model into the current frame.
Ci \etal~\cite{ci2018video} pre-scan the whole video and generate pseudo labels, then retrain the model based on these labels.
Some works ~\cite{azimi2022self,volpi2022road} discuss some existing image test-time training methods in handling challenging videos with human corruption.
A concurrent work~\cite{wang2023test} extends TTT-MAE to video streams. The model is initialized from the previous model and trained on a set of available frames.

\section{Preliminaries}
\label{Preliminary}

\noindent\textbf{Test-time Training.}
Test-time training~(TTT)~\cite{sun2020test,liu2021ttt++,gandelsman2022test} aims to adapt the pre-trained model to a new test distribution with a well-designed objective function without supervision.
A commonly used TTT network includes a shared encoder $\encoder$ and two decoders for the main task $\demain$ and self-supervised task $\dessl$. 
Typical TTT framework involves two-stage training. In the first stage, the network is trained with the main loss $\lossmain$ and self-supervised loss $\lossssl$: 
\begin{equation} \label{eqn:1st_stage}
 \min _{\encoder, \demain, \dessl} \lossmain + \sslweight \lossssl,
\end{equation}
\noindent where $\sslweight$ is a hyper-parameter to balance the two components.

In the second stage, which is referred as test-time training (TTT), for each individual input, the encoder~$\encoder$ will be fine-tuned according to the self-supervised objective: $\min _{\encoder} \lossssl$.

Another natural choice is to fine-tune also the $\dessl$. Empirically, the difference is negligible between only fine-tuning $\encoder$ and fine-tuning the extra $\dessl$~\cite{sun2020test}.

\noindent\textbf{Zero-shot Video Object Segmentation.}
Zero-shot video object segmentation~(ZSVOS) aims to localize the moving object in the video without any guidance during inference.
Current ZSVOS models~\cite{ren2021reciprocal,pei2022hierarchical,cho2023treating} consists of an image encoder for visual feature extraction, a flow encoder for motion information, and a decoder to get the mask prediction.
Given a video frame and its optical flow map, the image and flow encoder extract the multi-scale image and flow features, respectively. The aggregation of the image and flow features is used in the decoder to decode the object mask.
The ground truth object mask is used to supervise the model via the binary cross-entropy loss, which serves as the main loss $\lossmain$ if considering TTT on ZSVOS.

\begin{figure*}[t]
\centering
\includegraphics[width=\linewidth]{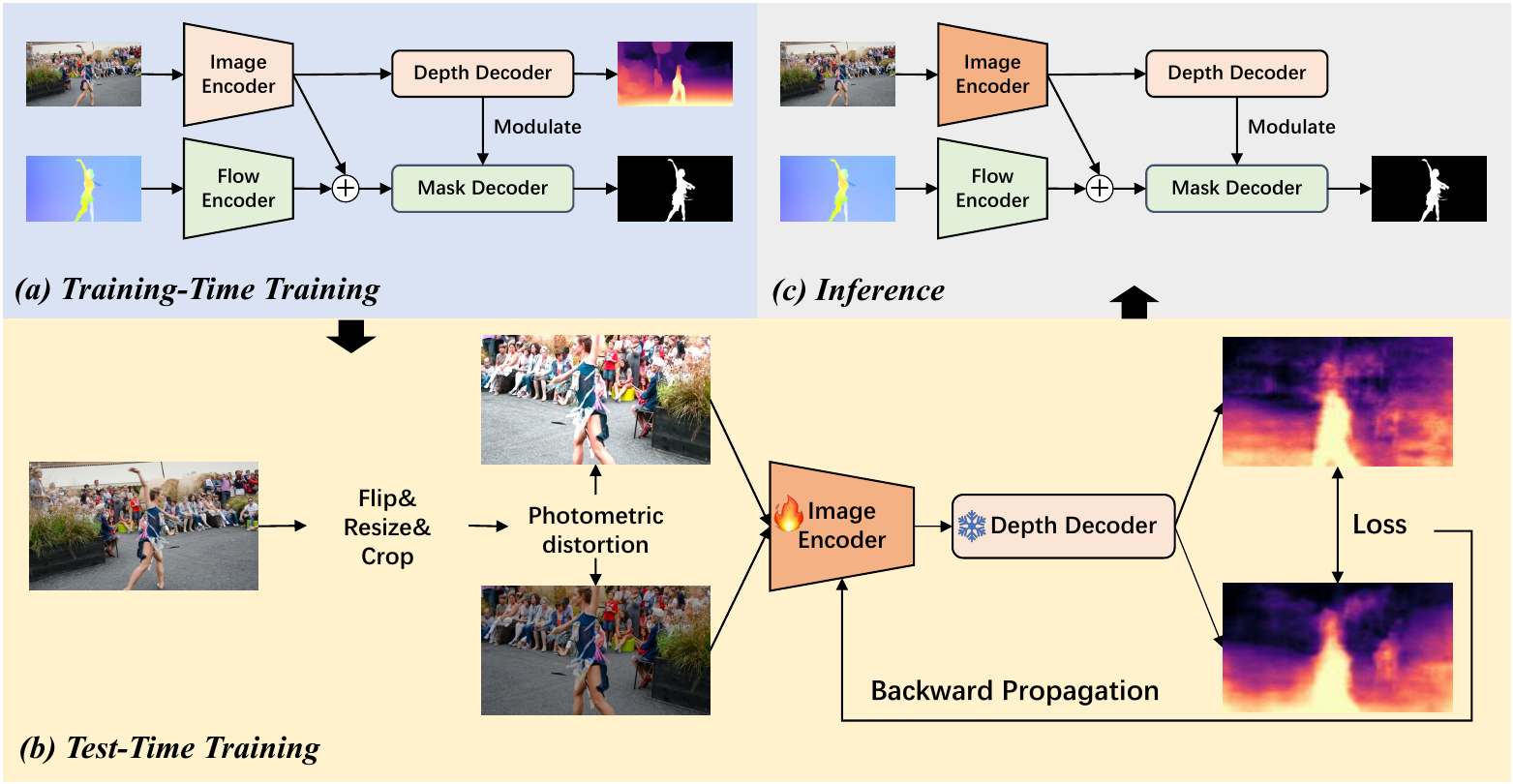}
\caption{\textbf{Overview of the proposed Depth-aware Test-Time Training.} We add a depth decoder to commonly used two-stream ZSVOS architecture to learn 3D knowledge. The model is first trained on large-scale datasets for object segmentation and depth estimation. Then, for each test video, we employ photometric distortion-based data augmentation to the frames. The error between the predicted depth maps is backward to update the image encoder. Finally, the new model is applied to infer the object. 
}
\label{fig:framework}
\end{figure*}

\section{Depth-aware Test-time Training for ZSVOS}
In this section, we present our Depth-aware Test-time Training (DATTT) for ZSVOS. 
Our entire framework is illustrated in Figure~\ref{fig:framework}. Our DATTT is designed following TTT-Rot~\cite{sun2020test} and TTT-MAE~\cite{gandelsman2022test}. During the first stage training, which is denoted as training-time training, DATTT is trained on a large-scale dataset to perform two tasks jointly: primary moving object segmentation and depth estimation, which are realized by the main task decoder $\demask$ and the depth decoder $\dedepth$ respectively. During the TTT, the model is required to produce consistent depth maps between two augmented samples. The error is used to update the image encoder $\enimage$ to better understand the current scene, thus is expected to give better mask prediction.

We start with training-time training in Section~\ref{sec:1st_stage_train} to introduce our first stage training. We then detail our TTT on videos in Section~\ref{sec:depth_aware_ttt}.

\subsection{Training-time Training}
\label{sec:1st_stage_train}

\noindent\textbf{Objective function.} Given an input frame $\inframe$ and flow $\inflow$, we first extract their features through an image encoder $\enimage$ and the flow encoder $\enflow$, as illustrated in Figure~\ref{fig:framework} (a). The encoded image and flow features are aggregated to predict the primary moving object through the mask decoder $\demask$. In this work, we leverage a simple summation as the aggregation of image and flow features.
The depth decoder $\dedepth$ estimates the depth map using the image features. We use the depth maps obtained by off-the-shelf monocular depth estimation methods~\cite{godard2019digging,zhang2023lite,bhat2023zoedepth} as pseudo ground truth $d$. The impact of different monocular depth estimation methods are provided in Section~\ref{sec:ana}.

Denoting the ground-truth mask as $m$. The total objective function can be formulated as:
\begin{equation} \label{eqn:loss}
 \lossbce(\demask(\enimage(\vidframe) + \enflow(\flow)),  )  +   \sslweight \lossdepth(\dedepth(\enimage(\vidframe)), d)
\end{equation}

\noindent where $\lambda$ is the hyper-parameter to balance the two losses. $\lossdepth$ is the standard scale-invariant log loss~\cite{eigen2014depth} for depth estimation following~\cite{bhat2023zoedepth}.

\begin{figure}[t]
\centering
\includegraphics[width=\linewidth]{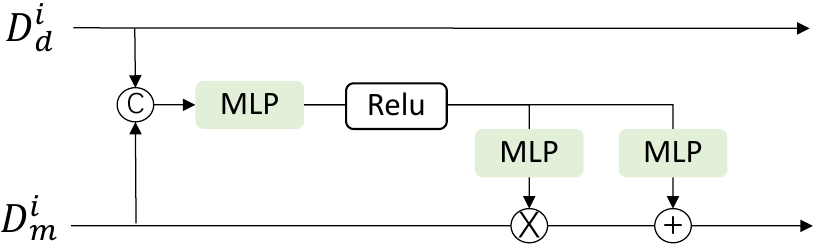}
\caption{
\textbf{The proposed depth-aware modulation layer}. At each scale $i$, we generate the modulation parameter by the depth feature $\dedepth^i$ and the object feature $\demask^i$ to modulate $\demask^i$.
}
\label{fig:modulation}
\end{figure}

\noindent\textbf{Depth-aware Modulation Layer.} We introduce the depth-aware modulation layer, which is illustrated in Figure~\ref{fig:modulation}. The basic idea is to enable features in the mask decoder to receive information from features in the depth decoder. 

Denoting the \textit{i}-th scale of the feature from the depth decoder and the mask decoder as $\dedepth^i$ and $\demask^i$ respectively, the depth-aware modulation layer is represented as $\modulation$ which serves to update $\demask^i$:
\begin{equation} \label{eqn:modulation}
 \demask^i =  \modulation(\demask^i, \dedepth^i)
\end{equation}
\noindent where $\modulation$ is composed of standard operators, such as MLP, Relu, concatenation, dot product, and summation, which can be seen in Figure~\ref{fig:modulation}.

\subsection{Test-time Training on Videos}
\label{sec:depth_aware_ttt}

\noindent\textbf{Depth-aware TTT.} Given a test video with $T$ frames $V=\left\{v_t | t \in [1, 2, ..., T] \right\}$, we conduct TTT to update the image encoder $\enimage$ by optimizing consistent depth map between a single frame under two data augmentation (Figure~\ref{fig:framework} (b)). The updated image encoder $\enimage$ is expected to be beneficial for the mask prediction (Figure~\ref{fig:framework} (c)).

Precisely, for the \textit{i}-th frame $v_i$, we obtain two augmented images $v_i^1$ and $v_i^2$ by applying different data augmentation on $v_i$, which includes: random horizontal flip, resize, crop, and photometric distortion. 
Then for TTT, we seek to optimize: 
\begin{equation} \label{eqn:loss_consistent}
 \lossdepth(\dedepth(\enimage(v_i^1)), \dedepth(\enimage(v_i^2)))
\end{equation}

Note that we keep the $\dedepth$ frozen and only fine-tune the image encoder $\enimage$, this is consistent with~\cite{sun2020test}. We also find that training $\dedepth$ and $\enimage$ together cannot bring improvement.

\begin{figure}[t]
\centering
\includegraphics[width=\linewidth]{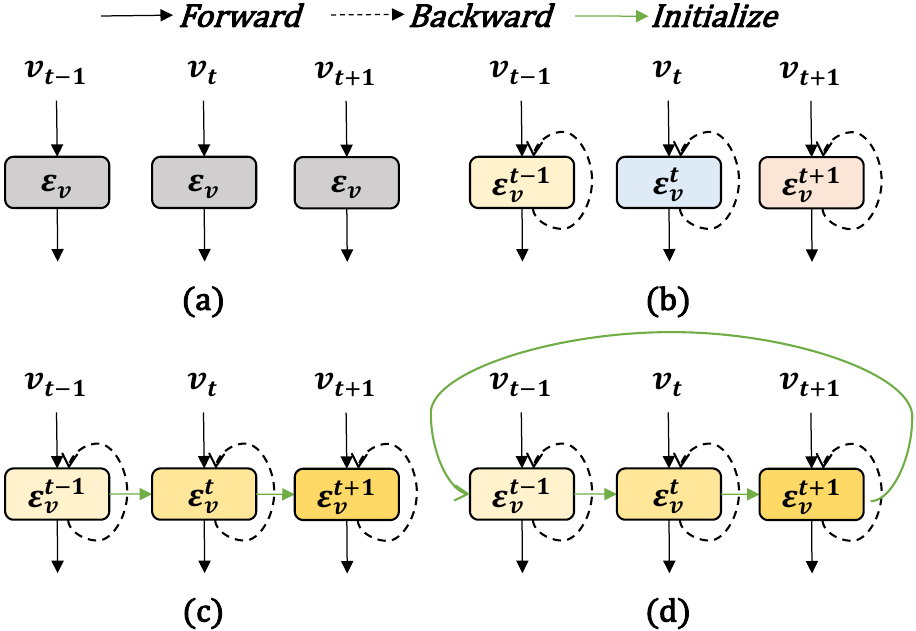}
\caption{
\textbf{A glance at different frameworks for ZSVOS described in Section~\ref{sec:depth_aware_ttt}.} (a)~The previous ZSVOS methods directly apply the trained model to infer the test video. (b)~Image-based test-time training methods~(TTT-N) fine-tune the model on each individual frame. (c)~Video test-time training by momentum-based weight initialization~(TTT-MWI) trains the model based on past models. (d)~Video test-time training by looping through the video~(TTT-LTV) benefits from the global information.
}
\label{fig:ttt_scheme}
\end{figure}

\noindent\textbf{Naive TTT strategy (TTT-N).} For each video, we train its own image encoder for each frame. This is a naive image test-time training strategy as Azimi~\etal~\cite{azimi2022self} (Figure~\ref{fig:ttt_scheme} (b)).
It treats the video as individual frames and adapts each frame by initializing the model with pre-trained weights. 
Although the model is adapted to each frame during testing, this strategy does not obtain additional benefits from the available frames in the video.

However, video is composed of a series of highly correlated images. The spatio-temporal correspondence in the video might boost the test-time training in the video data. Hence we introduce two effective strategies for video TTT: Momentum-based Weight Initialization (TTT-MWI) and Loop Through the Video (TTT-LTV). Note that the study of different strategies are provided in Section~\ref{sec:ana}.

\noindent\textbf{Momentum-based Weight Initialization (TTT-MWI).}
Since the scenes in consecutive frames in a video are highly similar, the model that has been finetuned on the past frame is more suitable to initialize than the pre-trained model when tuning in the current frame.
Therefore, we adapt the model to the video following the temporal order, where the parameters of the image encoder $\enimage^t$ for \textit{t}-th frame $v_t$ is initialized by the previous $\enimage^{t-1}$ instead of the original weight $\enimage$, which can be seen from Figure~\ref{fig:ttt_scheme} (c). 
In this way, the model is initialized by a better weight which has been adapted to the current scene based on past frames. The same strategy is also demonstrated to be efficient in the related work~\cite{volpi2022road}.
However, they only discuss TTT in online video streaming, here we further explore TTT in offline video.

\noindent\textbf{Loop Through the Video (TTT-LTV).}
The model is trained on each arrival frame for several epochs in the online setting. 
The model benefits from retaining information in past frames by momentum-based weight initialization since the past information is helpful.
In some offline settings~(such as video editing), the entire video is available.
To exploit more information in the entire video, we suggest performing video TTT by looping through the video rather than frame-by-frame.
Instead of training several epochs in the current frame and then moving to the next frame, the model is adapted at each frame once and loops for several epochs in the video (Figure~\ref{fig:ttt_scheme} (d)).
The scene knowledge is accumulated epoch-by-epoch and then serves as the past and future knowledge for the model for the current frame to train in a global view.

\section{Experiments}

\subsection{Datasets and Evaluation Metrics}
We evaluate the proposed method on five widely-used datasets, including DAVIS-16~\cite{perazzi2016benchmark}, FBMS~\cite{ochs2013segmentation}, 
Long-Videos~\cite{liang2020video}, MCL~\cite{kim2015spatiotemporal}, and SegTrackV2~\cite{li2013video}.
DAVIS-16~\cite{perazzi2016benchmark} contains a total of 50 videos with pixel-level annotations for each frame, including 30 videos for training and 20 videos for validation.
FBMS~\cite{ochs2013segmentation} consists of 29 training videos and 30 testing videos, with only 720 annotated frames.
Long-Videos~\cite{liang2020video} contains 3 long videos with each over 1,500 frames.
MCL~\cite{kim2015spatiotemporal} is composed of 9 videos in the low-resolution.
SegTrackV2~\cite{liang2020video} involves 14 videos of fast motion and object deformation.
Youtube-VOS~\cite{xu2018youtube} is used to train the model, which is a large-scale dataset including 3,471 videos.
Region similarity $\mathcal{J}$ and boundary accuracy $\mathcal{F}$ are reported for evaluation. 
$\mathcal{J}$ is defined as:
\begin{equation}
\mathcal{J}=\left|\frac{m_{gt} \cap m_{pred}}{m_{gt} \cup m_{pred}}\right|,
\end{equation}
where $m_{gt}$ and $m_{pred}$ are the ground truth mask and predicted mask, respectively.
$\mathcal{F}$ can be calculated as:
\begin{equation}
\mathcal{F}=\frac{2 \times p \times r}{p + r},
\end{equation}
where $p=\left|\frac{m_{gt} \cap m_{pred}}{m_{pred}}\right|$ and $r=\left|\frac{m_{gt} \cap m_{pred}}{m_{gt}}\right|$.

\subsection{Implementation Details}
All the experiments are performed on a single NVIDIA A40 GPU.
Random horizontal flipping, resizing, cropping, and photometric distortion are used for data augmentation. The input images are resized into $512 \times 512$.
The model is pre-trained on the Youtube-VOS dataset for 10 epochs with setting $\lambda = 0.1$. The ablation of $\lambda$ is provided 
in the appendix, Section~\ref{sec:lambda}.
During test-time training, we train the model for 10 epochs in each test video.
The mini-batch size is set to 8.  The model is optimized by the Adam optimizer with a learning rate of $6e^{-5}$ and $1e^{-5}$ for training-time training and test-time training.
We choose Mit-b1~\cite{xie2021segformer} and Swin-Tiny~\cite{liu2021swin} as the image encoder and flow encoder. The depth decoder and segmentation decoder are implemented with the lightweight decoder in SegFormer~\cite{xie2021segformer}.
RAFT~\cite{teed2020raft} is used to extract the optical flow map.
Monodepth2~\cite{godard2019digging}, LiteMono~\cite{zhang2023lite}, and ZoeDepth~\cite{bhat2023zoedepth} are used to obtain the depth map.
We use Mit-b1 as the backbone, MonoDepth2 as the depth extractor, and TTT-LTV as the TTT strategy in our default setting.

\begin{table}[!t]
\centering
\resizebox{\linewidth}{!}
{
\begin{tabular}{l|lcc|ccc}
\toprule
Backbone & $\dedepth$ & Mod.  & TTT & DAVIS-16 & FBMS & Long. \\\hline

\multirow{5}{*}{ \shortstack{Mit-b1\\~\cite{xie2021segformer}}}    
& -   & -   & -   & 75.9  & 75.1 & 63.9  \\  \cline{2-7} 
&  \checkmark & -  & - & 77.0  & 77.5 & 62.8 \\ 
&  \checkmark & -  & \checkmark & 77.2  & \textbf{78.0} & 70.5 \\ \cline{2-7} 
&  \checkmark & \checkmark & -    & 77.1 & 73.7 & 65.2  \\ 
&  \checkmark & \checkmark  & \checkmark  & \textbf{77.5} & 76.9 & \textbf{73.1}  \\ \hline\hline

\multirow{5}{*}{ \shortstack{Swin-T\\~\cite{liu2021swin}}}    
& -   & -   & -  & 77.8  & 74.1 & 65.7 \\ \cline{2-7} 
&  \checkmark & -  & -   & 78.7  & 74.5 & 67.0  \\ 
&  \checkmark & -  & \checkmark & 78.8  & 75.0 & 72.1  \\ \cline{2-7} 
&  \checkmark & \checkmark & -  & 79.0  &  76.6 & 63.5 \\ 
&  \checkmark & \checkmark  & \checkmark & \textbf{79.2}  & \textbf{79.2} & \textbf{75.9} \\ 

\hline
\end{tabular}}
\vspace{-1mm}
\caption{\textbf{Ablation study for the proposed depth-aware decoder on DAVIS-16~\cite{perazzi2016benchmark}, FBMS~\cite{ochs2013segmentation}, and Long-Videos~\cite{liang2020video} datasets.} $\mathcal{J}$ is reported for comparison. Taking depth as additional supervision~($\dedepth$) boosts performance, and the modulation layer~(Mod.) obtains a more significant improvement during TTT.
}
\label{tab:ablation_arch}
\end{table}

\begin{table}[!t]
\centering
\resizebox{\linewidth}{!}
{
\begin{tabular}{l|c|ccc}
\toprule
Depth Extractors & TTT & DAVIS-16 & FBMS & Long. \\\hline
w/o depth & - & 75.9  & 75.1 & 63.9  \\ \hline\hline
\multirow{2}{*}{ \shortstack{MonoDepth2~\cite{godard2019digging}}}   & -   & 77.1 & 73.7 & 65.2   \\ 
 & \checkmark & \footnotesize$+0.4$  & \footnotesize$+3.2$ & \footnotesize$+7.9$ \\ \hline\hline
\multirow{2}{*}{ \shortstack{LiteMono~\cite{zhang2023lite}}}  & -   & 76.8  & 79.0 & 68.1  \\ 
 & \checkmark & \footnotesize$+2.0$  & \footnotesize$+1.5$ & \footnotesize$+6.3$  \\ \hline\hline
\multirow{2}{*}{ \shortstack{ZoeDepth~\cite{bhat2023zoedepth}}}  & -    & 79.9 & 76.4 & 64.0 \\ 
 & \checkmark  & \footnotesize$+0.5$ & \footnotesize$+4.7$ & \footnotesize$+9.5$  \\ 
\bottomrule
\end{tabular}}
\caption{\textbf{Ablation study using different depth estimation methods on DAVIS-16~\cite{perazzi2016benchmark}, FBMS~\cite{ochs2013segmentation}, and Long-Videos~\cite{liang2020video} datasets.} $\mathcal{J}$ is reported for comparison. DATTT shows consistent improvements using different depth estimation methods.}
\vspace{-1em}
\label{tab:ablation_depth_method}
\end{table}

\begin{table*}[!t]
\centering
{
\begin{tabular}{l|l|cc|cc|cc|cc|cc}
\toprule
 \multirow{2}{*} {Backbone} & \multirow{2}{*} {TTT Scheme} & \multicolumn{2}{c|}{DAVIS-16} & \multicolumn{2}{c|}{FBMS} & \multicolumn{2}{c|}{Long.} & \multicolumn{2}{c|}{MCL} & \multicolumn{2}{c}{STV2} \\ 
 & &  $\mathcal{J}$ & $\mathcal{F}$ & $\mathcal{J}$ & $\mathcal{F}$ & $\mathcal{J}$ & $\mathcal{F}$ &$\mathcal{J}$ & $\mathcal{F}$ &$\mathcal{J}$ & $\mathcal{F}$ \\ \hline
\multirow{4}{*}{ \shortstack{Mit-b1~\cite{xie2021segformer}}}
& -  & 77.1 & 78.4 & 73.7 & 75.8 & 65.2 & 68.0 & 53.5 & 66.2 & 61.5 & 69.2 \\
& TTT-N  & \footnotesize$+0.3$ & \footnotesize$+0.3$ & \footnotesize$+0.1$ & \footnotesize$+0.3$ & \footnotesize$+1.3$ & \footnotesize$+1.5$ & \footnotesize$+1.9$ & \footnotesize$+1.5$ & \footnotesize$+1.0$ & \footnotesize$+1.2$ \\
& TTT-MWI & \footnotesize$+0.4$ & \footnotesize$+\textbf{0.5}$ & \footnotesize$+2.3$ & \footnotesize$+2.1$ & \footnotesize$+7.2$ & \footnotesize$+7.5$ & \footnotesize$+7.6$ & \footnotesize$+6.9$ & \footnotesize$+2.0$ & \footnotesize$+4.0$ \\
& TTT-LTV & \footnotesize$+\textbf{0.4}$ & \footnotesize$+0.4$ & \footnotesize$+\textbf{3.2}$ & \footnotesize$+\textbf{3.1}$ & \footnotesize$+\textbf{7.9}$ & \footnotesize$+\textbf{7.7}$ & \footnotesize$+\textbf{8.4}$ & \footnotesize$+\textbf{7.8}$ & \footnotesize$+\textbf{4.4}$ & \footnotesize$+\textbf{4.3}$ \\ \hline \hline

\multirow{4}{*}{ \shortstack{Swin-T~\cite{liu2021swin}}}
& -  & 79.0 & 80.3 & 76.6 & 79.3 & 63.5 & 70.0 & 54.1 & 68.2 & 64.0 & 70.7\\
& TTT-N & \footnotesize$+0.1$ & \footnotesize$+0.2$ & \footnotesize$+1.2$ & \footnotesize$+1.0$ & \footnotesize$+2.6$ & \footnotesize$+1.7$ & \footnotesize$+1.3$ & \footnotesize$+1.4$ & \footnotesize$+0.4$ & \footnotesize$+0.4$ \\
& TTT-MWI & \footnotesize$+\textbf{0.3}$ & \footnotesize$+0.4$ & \footnotesize$+2.2$ & \footnotesize$+1.7$ & \footnotesize$+7.6$ & \footnotesize$+5.7$ & \footnotesize$+8.1$ & \footnotesize$+6.0$ & \footnotesize$+1.4$ & \footnotesize$+0.6$ \\
& TTT-LTV & \footnotesize$+0.2$ & \footnotesize$+\textbf{0.4}$ & \footnotesize$+\textbf{2.6}$ & \footnotesize$+\textbf{2.0}$ & \footnotesize$+\textbf{12.4}$ & \footnotesize$+\textbf{9.2}$ & \footnotesize$+\textbf{12.0}$ & \footnotesize$+\textbf{8.1}$ & \footnotesize$+\textbf{1.5}$ & \footnotesize$+\textbf{0.8}$ \\ \bottomrule
\end{tabular}}
\vspace{-2mm}
\caption{\textbf{Ablation study on the proposed test-time training scheme on DAVIS-16~\cite{perazzi2016benchmark}, FBMS~\cite{ochs2013segmentation}, Long-Videos~\cite{liang2020video}, MCL~\cite{kim2015spatiotemporal}, and SegTrackV2~\cite{li2013video} datasets.}
The proposed strategy is effective for test-time training in video.}
\label{tab:ablation_ttt}
\end{table*}

\newcommand\ww{0.33\textwidth}
\newcommand\hh{0.33\textwidth}
\begin{figure*}[th]
\centering  

\includegraphics[width=\ww]{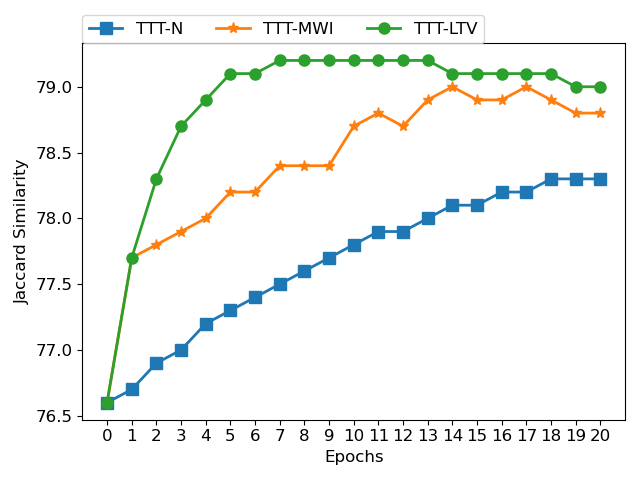}
\includegraphics[width=\ww]{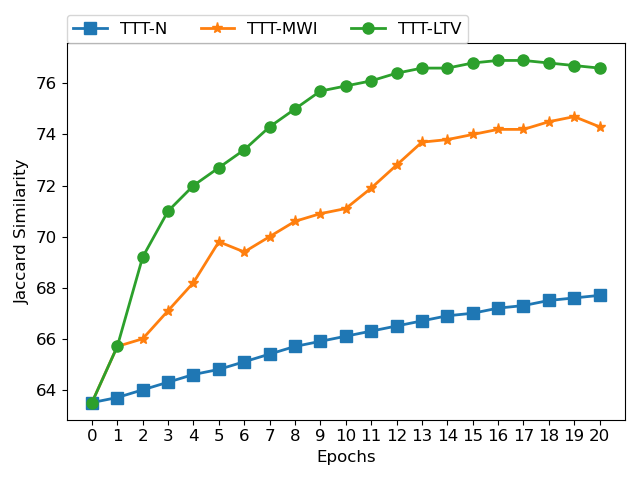}
\includegraphics[width=\ww]{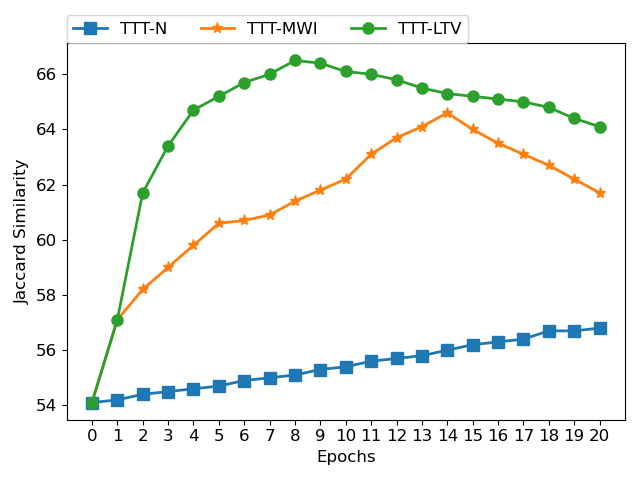}
\vspace{-1em}
\caption{\textbf{The performance varies with the number of training epochs on FBMS~\cite{ochs2013segmentation}, Long-Videos~\cite{liang2020video}, MCL~\cite{kim2015spatiotemporal} datasets.} The proposed strategy (TTT-LTV introduced in Section~\ref{sec:depth_aware_ttt}) requires less time for the model to adapt to the target video on the three datasets and achieves better results.}
\label{fig:ttt_epoch}
\end{figure*}

\subsection{Analysis and Ablation Studies}
\label{sec:ana}

\noindent\textbf{Impact of Architecture Design and Depth Quality.}
We first verify the proposed depth-aware decoder. The baseline is set to the commonly used two-stream model. As shown in Table~\ref{tab:ablation_arch}, our method is better than the baseline under two different backbones. The results are intuitive since depth information is beneficial to segment the primary object.
In addition, the depth modulation layer is more effective in both training-time training and test-time training. 
This can be attributed to the updated depth features further facilitating object segmentation via feature modulation in the decoders.

We also experiment with the depth maps obtained by different depth estimation methods as supervision. Table~\ref{tab:ablation_depth_method} shows that the proposed DATTT gains consistent improvement across different methods. 
It demonstrates that utilizing 3D information in the given video to finetune the model is effective for ZSVOS.

\noindent\textbf{Test-Time Training Strategy.}
We discuss different test-time training schemes proposed in Section~\ref{sec:depth_aware_ttt}. As shown in Table~\ref{tab:ablation_ttt}, they obtain consistent improvement in several datasets compared to directly testing on the video, which proves that the depth-aware test-time training is effective in ZSVOS.
However, the performances are significantly varying in different strategies. First, treating the video as a whole rather than individual frames greatly improves the performance of video test-time training. Utilizing the parameters of the previous frame without resetting the parameter to the pre-trained keeps the model remembering the past scenes. It becomes easier to find the primary object because of the temporal smoothness.
Additionally, it is useful to make the model iteratively train on the video in the offline setting. As the iteration goes on, both past and future information are available. The frames in the front of the video can be further refined after snooping future information.

\noindent\textbf{Training Epoch.}
One crucial challenge in TTT is the additional time required for training. We investigate the impact of training epochs on performance. The results are presented in Figure~\ref{fig:ttt_epoch}. As the number of training epochs increases, the disparity between TTT-LTV and other schemes~(TTT-N and TTT-MWI) becomes pronounced, indicating that the proposed approach requires less time to adapt to a given video.
Furthermore, we observe that the optimal number of training epochs varies for different datasets. However, a suitable epoch~(such as 10) yields satisfactory results across diverse datasets. We present the results obtained by our method in a video sequence in Figure~\ref{fig:visualization}. Note that more visual results are provided 
in the appendix, Figure~\ref{fig:supp_visualization}. 
Initially, the pre-trained model shows limited accuracy in detecting the walking people in the video. As we apply depth-based TTT, we observe a progressive improvement in the subsequent results.

\begin{figure*}[!th]
\centering
\includegraphics[width=\linewidth]{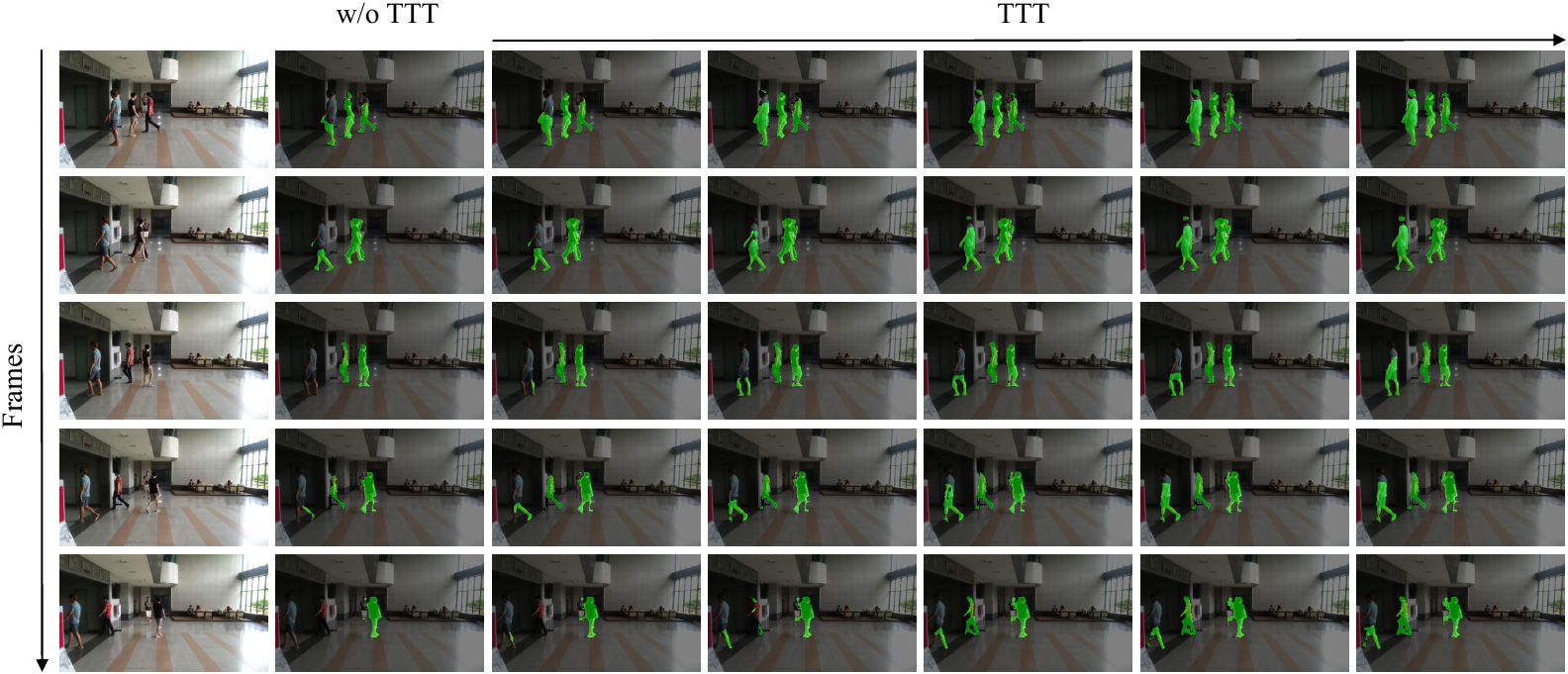}
\vspace{-2em}
\caption{\textbf{Qualitative results of the proposed method.} The background in the results is dimmed for better visualization. The results obtained by the pre-trained model are less accurate and become better and better as TTT goes on.}
\label{fig:visualization}
\end{figure*}

\begin{table*}[!t]
\centering
{
\begin{tabular}{l|l|l|ll|ll|ll|ll|ll}
\toprule
 \multirow{2}{*} {Backbone} & \multirow{2}{*} {Method} & \multirow{2}{*} {TTT} & \multicolumn{2}{c|}{DAVIS-16} & \multicolumn{2}{c|}{FBMS} & \multicolumn{2}{c|}{Long.} & \multicolumn{2}{c|}{MCL} & \multicolumn{2}{c}{STV2} \\ 
 & & & $\mathcal{J}$ & $\mathcal{F}$ & $\mathcal{J}$ & $\mathcal{F}$ & $\mathcal{J}$ & $\mathcal{F}$ & $\mathcal{J}$ & $\mathcal{F}$  & $\mathcal{J}$ & $\mathcal{F}$  \\ \hline
\multirow{9}{*}{ \shortstack{Mit-b1~\cite{xie2021segformer}}}
& Baseline  & - & 75.9 & 77.5 & 75.1 & 76.5 & 63.9 & 67.5 & 57.3 & 70.8 & 61.5 & 70.4 \\ 
& TENT~\cite{wang2020tent}  & \checkmark & \color{red}{\footnotesize$-0.5$} & \color{red}{\footnotesize$-0.4$} & \color{black}{\footnotesize$+0.4$} & \color{black}{\footnotesize$+0.4$} & \color{black}{\footnotesize$+0.6$} & \color{black}{\footnotesize$+0.5$} & \color{black}{\footnotesize$+1.0$} & \color{black}{\footnotesize$+0.4$} & \color{red}{\footnotesize$-0.3$} & \color{red}{\footnotesize$-0.4$} \\
& BN~\cite{schneider2020improving}  & \checkmark & \color{black}{\footnotesize$+0.3$} & \color{black}{\footnotesize$+0.3$} & \color{black}{\footnotesize$+0.9$} & \color{black}{\footnotesize$+1.0$} & \color{black}{\footnotesize$+0.9$} & \color{black}{\footnotesize$+0.6$} & \color{black}{\footnotesize$+1.4$} & \color{black}{\footnotesize$+1.3$} & \color{black}{\footnotesize$+0.3$} & \color{black}{\footnotesize$+0.3$} \\ \cline{2-13}
& TTT-Rot~\cite{sun2020test}  & - & 75.3 & 76.2 & 75.4 & 77.2 & 59.1 & 62.8 & 57.7 & 70.5 & 66.4 & 73.3 \\
&  & \checkmark & \color{red}{\footnotesize$-0.4$} & \color{red}{\footnotesize$-0.1$} & \color{black}{\footnotesize$+0.4$} & \color{black}{\footnotesize$+0.5$} & \color{black}{\footnotesize$+2.6$} & \color{black}{\footnotesize$+1.7$} & \color{black}{\footnotesize$+4.0$} & \color{black}{\footnotesize$+3.8$} & \color{red}{\footnotesize$-2.4$} & \color{red}{\footnotesize$-1.6$} \\ \cline{2-13}
& TTT-MAE~\cite{gandelsman2022test} & - & 73.5 & 74.1 & 74.6 & 75.7 & 64.4 & 67.5 & 55.7 & 66.8 & 62.2 & 70.2 \\
&  & \checkmark &  \color{black}{\footnotesize$+0.4$} & \color{black}{\footnotesize$+0.3$} & \color{black}{\footnotesize$+0.5$} & \color{black}{\footnotesize$+0.1$} & \color{black}{\footnotesize$+0.9$} & \color{black}{\footnotesize$+0.1$} & \color{red}{\footnotesize$-1.5$} & \color{red}{\footnotesize$-0.9$} & \color{red}{\footnotesize$-0.7$} & \color{red}{\footnotesize$-0.4$} \\ \cline{2-13}
& Ours  & - & 77.1 & 78.4 & 73.7 & 75.8 & 65.2 & 68.0 & 53.5 & 66.2 & 61.5 & 69.2 \\ 
&   & \checkmark & \color{black}{\footnotesize$+\textbf{0.4}$} & \color{black}{\footnotesize$+\textbf{0.4}$} & \color{black}{\footnotesize$+\textbf{3.2}$} & \color{black}{\footnotesize$+\textbf{3.1}$} & \color{black}{\footnotesize$+\textbf{7.9}$} & \color{black}{\footnotesize$+\textbf{7.7}$} & \color{black}{\footnotesize$+\textbf{8.4}$} & \color{black}{\footnotesize$+\textbf{7.8}$} & \color{black}{\footnotesize$+\textbf{4.4}$} & \color{black}{\footnotesize$+\textbf{4.3}$} \\ \hline  \hline

\multirow{9}{*}{ \shortstack{Swin-T~\cite{liu2021swin}}}
& Baseline  & - & 77.8 & 79.0 & 74.1 & 77.7 & 65.7 & 71.2 & 47.3 & 61.2 & 62.6 & 70.2 \\ 
& TENT~\cite{wang2020tent}  & \checkmark & \color{red}{\footnotesize$-0.4$} & \color{red}{\footnotesize$-0.5$} & \color{black}{\footnotesize$+0.3$} & \color{black}{\footnotesize$+0.3$}  & \color{black}{\footnotesize$+0.3$} & \color{black}{\footnotesize$+0.3$} & \color{black}{\footnotesize$+0.6$} & \color{black}{\footnotesize$+0.5$} & \color{red}{\footnotesize$-0.5$} & \color{red}{\footnotesize$-0.7$} \\
& BN~\cite{schneider2020improving}  & \checkmark & \color{red}{\footnotesize$-0.2$} & \color{red}{\footnotesize$-0.3$} & \color{black}{\footnotesize$+0.6$} & \color{black}{\footnotesize$+0.7$} & \color{black}{\footnotesize$+0.6$} & \color{black}{\footnotesize$+0.6$} & \color{black}{\footnotesize$+1.0$} & \color{black}{\footnotesize$+0.9$} & \color{black}{\footnotesize$+0.4$} & \color{black}{\footnotesize$+0.2$} \\ \cline{2-13}
& TTT-Rot~\cite{sun2020test}  & - & 78.9 & 79.9 & 75.5 & 78.5 & 66.7 & 70.6 & 57.0 & 69.7 & 63.3 & 70.4 \\
&   & \checkmark & \color{black}{\footnotesize$+\textbf{0.7}$} & \color{black}{\footnotesize$+\textbf{0.6}$} & \color{red}{\footnotesize$-1.0$} & \color{red}{\footnotesize$-0.1$} & \color{red}{\footnotesize$-1.8$} & \color{red}{\footnotesize$-2.3$} & \color{black}{\footnotesize$+0.6$} & \color{black}{\footnotesize$+0.4$} & \color{black}{\footnotesize$+0.4$} & \color{black}{\footnotesize$+0.3$} \\ \cline{2-13}
& TTT-MAE~\cite{gandelsman2022test} & - & 77.0 & 77.9 & 74.6 & 77.2 & 65.3 & 69.2 & 52.8 & 65.5 & 60.3 & 67.9 \\
&   & \checkmark & \color{red}{\footnotesize$-0.1$} &  \color{red}{\footnotesize$-0.1$} & \color{red}{\footnotesize$-0.3$} & \color{red}{\footnotesize$-0.1$} & \color{red}{\footnotesize$-0.8$} & \color{red}{\footnotesize$-0.6$} & \color{red}{\footnotesize$-0.9$} & \color{red}{\footnotesize$-0.8$} & \color{red}{\footnotesize$-0.6$} & \color{red}{\footnotesize$-0.4$} \\ \cline{2-13}
& Ours  & - & 79.0 & 80.3 & 76.6 & 79.3 & 63.5 & 70.0 & 54.1 & 68.2 & 64.0 & 70.7 \\ 
&   & \checkmark & \color{black}{\footnotesize$+0.2$} & \color{black}{\footnotesize$+0.4$} & \color{black}{\footnotesize$+\textbf{2.6}$} & \color{black}{\footnotesize$+\textbf{2.0}$} & \color{black}{\footnotesize$+\textbf{12.4}$} & \color{black}{\footnotesize$+\textbf{9.2}$} & \color{black}{\footnotesize$+\textbf{12.0}$} & \color{black}{\footnotesize$+\textbf{8.1}$} & \color{black}{\footnotesize$+\textbf{1.5}$} & \color{black}{\footnotesize$+\textbf{0.8}$} \\ \hline 
\end{tabular}}
\vspace{-2mm}
\caption{\textbf{Comparisons with state-of-the-art test-time training method on DAVIS-16~\cite{perazzi2016benchmark}, FBMS~\cite{ochs2013segmentation}, Long-Videos~\cite{liang2020video}, MCL~\cite{kim2015spatiotemporal}, and SegTrackV2~\cite{li2013video} datasets.}  Results that 
the dropped after TTT are masked as {\color{red}{red}}. The most significant improvement is marked as \textbf{bold}. The proposed method obtains stable improvements in diverse datasets. 
}
\label{tab:sota_ttt}
\end{table*}

\newcommand{\ablationtablestyle}[2]{\setlength{\tabcolsep}{#1}\renewcommand{\arraystretch}{#2}\centering\small}

\begin{table*}[t]
\centering
\ablationtablestyle{2pt}{1.}
{
\begin{tabular}{c|c|cccccc|cc}
\toprule
&  & \small{3DCSEG~\cite{mahadevan2020making}} & \small{AGNN~\cite{wang2019zero}} & \small{MATNet~\cite{zhou2020motion}}  & \small{HFAN~\cite{pei2022hierarchical}} & \small{HCPN~\cite{pei2023hierarchical}}  & \small{MED-VT~\cite{karim2023med}} & \small{Ours} & \small{Ours} \\ 
& & \footnotesize{3D ResNet-152~\cite{hara2018can}} & \footnotesize{Resnet-101~\cite{he2016deep}} & \footnotesize{Resnet-101~\cite{he2016deep}} & \footnotesize{Mit-b1~\cite{xie2021segformer}} & \footnotesize{Resnet-101~\cite{he2016deep}} & \footnotesize{Video-Swin-B~\cite{liu2022video}} & \footnotesize{Mit-b1~\cite{xie2021segformer}}& \footnotesize{Swin-T~\cite{liu2021swin}}\\ \hline

\multirow{2}{*}{ \shortstack{DAVIS-16}} & ~$\mathcal{J}$~ & 84.3 & 80.7 & 82.4  & \textbf{86.2} & 85.8    & 85.9 & 86.0 & 85.8 \\
& $\mathcal{F}$  & 84.7 & 79.1 & 80.7  & 87.1 & 85.4 & 86.6  & 87.9 & \textbf{88.5} \\ \hline
FBMS & $\mathcal{J}$ & 76.2  & - & 76.1  & 76.1 & 78.3   & -  &  74.9 & \textbf{78.8} \\ \hline
\multirow{2}{*}{ \shortstack{Long.}} & $\mathcal{J}$ & 34.2  & 68.3 & 66.4  & 74.9 & -  & - & 75.6 & \textbf{77.3} \\
& $\mathcal{F}$  & 33.1 & 68.6 & 69.3   & 76.1 & -  & -  & 77.1 & \textbf{79.9} \\
\bottomrule
\end{tabular}
}
\vspace{-2mm}
\caption{\textbf{Comparison with state-of-the-art ZSVOS methods on DAVIS-16~\cite{perazzi2016benchmark}, FBMS~\cite{ochs2013segmentation}, and Long-Videos~\cite{liang2020video} datasets.} The proposed method is superior to other ZSVOS methods that directly test on videos. 
}
\label{tab:sota_zsvos}
\end{table*}

\subsection{Comparison with the State-of-the-Art}
We first compare with four prior methods for test-time training~(TTT) in Table~\ref{tab:sota_ttt}, including TENT~\cite{wang2020tent}, BN~\cite{schneider2020improving}, TTT-Rot~\cite{sun2020test}, and TTT-MAE~\cite{gandelsman2022test}.
In detail, we directly apply TTT to the baseline model for methods that do not incorporate an auxiliary head~(TENT and BN). For methods that employ an auxiliary head~(TTT-Rot, TTT-MAE, and ours), we train the respective baseline models and then perform TTT.  All these methods follow the TTT strategy we have designed as described above.

It is observed that these methods do not yield substantial improvements in ZSVOS. The prior work~\cite{volpi2022road} gets similar results, where these methods demonstrate promising performance in challenging weather conditions like rain or snow but fail to produce significant improvements in typical weather scenarios such as sunny. According to their analysis, the baseline model has achieved satisfactory performance by training on typical weather data, thereby limiting its potential for further enhancements.
Instead, we attribute this limitation to no synthetic corruption in the target datasets, resulting in a minor domain shift between the training and test data. Severe domain shifts make the model easy to adapt to new samples via even solely updating the normalization statistics. Without clear domain shifts, the presence of task-specific cues becomes crucial to effectively guide TTT. In the context of ZSVOS, cues derived from 3D information prove to be advantageous, which explains the stable improvements exhibited by our proposed method when applied to test data.

Then, in auxiliary head-based TTT approaches, adding auxiliary tasks can not improve the baseline performance consistently. For instance, when the shared image encoder is required to learn high-level semantic representations through image reconstruction, it frequently undermines its low-level features that are essential for object segmentation. The success observed in rotation prediction can be attributed to the additional data augmentation introduced by varying degrees of rotation. These auxiliary tasks both exhibit limited impact when employed to TTT for ZSVOS, while our depth-based method shows notable improvements.

We also compared the proposed method with SOTA ZSVOS methods~\cite{mahadevan2020making,wang2019zero,zhou2020motion,pei2022hierarchical,pei2023hierarchical,karim2023med}. These methods are usually pre-trained on large-scale datasets and then fine-tuned on the DAVIS-16 training set. For a fair comparison, we finetune the model that has been pre-trained on Youtube-VOS on the DAVIS-16 training set and perform TTT on the FBMS and Long-Videos dataset. As shown in Table~\ref{tab:sota_zsvos}, our method obtains the best performance in 2 of 3 datasets. 
Note that our method involves no bells and whistles blocks for feature fusion, information propagation, attention mechanism, \etc, suggesting performing TTT is effective for ZSVOS.

\section{Conclusion}
In this work, we introduce Depth-aware Test-Time Training for ZSVOS, which allows a pre-trained model better generalize to unseen scenarios.
We propose a joint learning framework that simultaneously addresses object segmentation and depth estimation.
During the inference, the consistent depth for the same frame under different data augmentations serves as the criterion for updating the model. Furthermore, different TTT strategies are explored. The experimental results demonstrate the effectiveness of our proposed approach in comparison to SOTA TTT approaches. We also achieve competitive performance compared to other ZSVOS methods.

{\textbf{Acknowledgments:} This work was supported in part by the Science and Technology Development Fund, Macau SAR,
under Grant 0087/2020/A2 and 0141/2023/RIA2, in part by the National Natural Science Foundation of China under Grant 62376046.}

{
    \small
    \bibliographystyle{ieeenat_fullname}
    \bibliography{main}
}


\clearpage

\appendix
\label{sec:appendix}
\vspace*{1em}{\centering\Large\bf%
Appendix
\vspace*{1.5em}}

\renewcommand\thetable{\Alph{table}}   
\renewcommand\thefigure{\Alph{figure}}    
\renewcommand\thesection{\Alph{section}} 

\definecolor{cvprblue}{rgb}{0.21,0.49,0.74}
This appendix 
contains the following sections: 
\begin{itemize}
\item Section~\ref{sec:efficiency}: Efficiency analysis.
\item Section~\ref{sec:data_aug}: Impact of different data augmentation for TTT~(\textit{c.f.} Section~\ref{sec:depth_aware_ttt}).
\item Section~\ref{sec:pseudo_depth}: TTT with depth supervision from depth predictor.
\item Section~\ref{sec:tttn}: Further study on Naive TTT (TTT-N) strategy~(\textit{c.f.} Section~\ref{sec:depth_aware_ttt}).
\item Section~\ref{sec:lambda}: Impact of the depth prediction loss weight $\lambda$~(\textit{c.f.} Section~\ref{sec:1st_stage_train}).
\item Section~\ref{sec:sampling}: Sampling strategy for densely annotated videos: DAVIS-16~\cite{perazzi2016benchmark} and SegTrackV2~\cite{li2013video}. 
\item Section~\ref{sec:extend}: Extending DATTT to existing ZSVOS method. 
\item Section~\ref{sec:vis}: Additional visual results~(\textit{c.f.} Figure~\ref{fig:visualization}).
\end{itemize}

\section{Efficiency analysis}
\label{sec:efficiency}
When utilizing Mit-b1~\cite{xie2021segformer} as our backbone, the parameters in each component are distributed as follows: 26.30M for the backbone, 0.53M for the segmentation head, 0.53M for the depth head, and 1.05M for the depth-aware modulation layer.  Consequently, we introduce only a minimal number of additional parameters to the baseline model.  The baseline model, consisting of the backbone with the segmentation head, requires 1.4 hours per epoch for training and 8 milliseconds per frame for inference.  In contrast, our full model, which includes the baseline, the depth head, and the depth-aware modulation layer, demands 1.5 hours per epoch for training, 9 milliseconds per frame for inference, and 49 milliseconds per frame for TTT.  The power consumption on the GPU is recorded as 230W for training, 80W for inference, and 200W for TTT, indicating that the additional computation cost to the basic training and inference processes is minimal.  While the incorporation of TTT adds to the computational time, it is an inherent drawback of the technique, and we strive to mitigate its impact. It is worth noting that our proposed strategy is significantly faster and more effective than the naive strategy~(\textit{c.f.} Figure~\ref{fig:ttt_epoch}).

\section{Impact of different data augmentation for TTT}
\label{sec:data_aug}
We apply random horizontal flipping, resizing, cropping, and photometric distortion for data augmentation. 
In detail, photometric distortion includes random brightness, contrast, saturation, and hue.
We further ablate the effect of each type of augmentation on it. 
Table~\ref{tab:supp_data_aug} shows that the proposed method works well when removing any one kind of data augmentation, which indicates that our success does not come from any particular trick.
Each type of augmentation doesn't affect the model significantly since it is used to create a pair of samples for consistent depth map optimization. Therefore, the proposed TTT strategy is the key to success.

\begin{table}[!t]
\centering
{
\begin{tabular}{l|ccc}
\toprule
 & DAVIS-16 & FBMS & Long. \\\hline
- & 77.1  & 73.7 & 65.2  \\ \hline\hline

w/o resize & \footnotesize$+0.5$  & \footnotesize$+3.2$ & \footnotesize$+7.7$ \\ 

w/o crop & \footnotesize$+0.4$  & \footnotesize$+2.5$ & \footnotesize$+7.5$  \\ 
 
w/o flip & \footnotesize$+0.4$ & \footnotesize$+3.1$ & \footnotesize$+7.2$  \\ 
\hline \hline


w/o brightness & \footnotesize$+0.4$  & \footnotesize$+2.9$ & \footnotesize$+7.3$ \\ 
w/o contrast & \footnotesize$+0.7$  & \footnotesize$+2.8$ & \footnotesize$+7.3$ \\ 
w/o saturation & \footnotesize$+0.3$  & \footnotesize$+2.9$ & \footnotesize$+7.7$ \\ 
w/o hue & \footnotesize$+0.4$  & \footnotesize$+2.8$ & \footnotesize$+6.4$ \\ 
\hline \hline
full & \footnotesize$+0.4$ & \footnotesize$+3.2$ & \footnotesize$+7.9$  \\ 
\bottomrule
\end{tabular}}
\caption{\textbf{Impact of different data augmentation for TTT in DAVIS-16~\cite{perazzi2016benchmark}, FBMS~\cite{ochs2013segmentation}, 
Long-Videos~\cite{liang2020video} datasets.} $\mathcal{J}$ is reported for comparison.}
\label{tab:supp_data_aug}
\end{table}

\section{TTT with depth supervision from depth predictor}
\label{sec:pseudo_depth}
We also experiment with pre-calculated depth maps for test-time training. 
This means that instead of generating two batches of images to minimize the distance between their depth maps (`\textit{Consistent Depth}` in Table~\ref{tab:supp_loss}), we predict one batch of depth maps and calculate the error with the pseudo depth maps (`\textit{Pseudo Depth}` in Table~\ref{tab:supp_loss}). This process is similar to the training-time training stage~(\textit{c.f.} Section~\ref{sec:1st_stage_train}), but without the binary cross entropy for segmentation. As shown in Table~\ref{tab:supp_loss}, it brings less improvement and sometimes fails. This can be due to the model being explicitly required to learn depth estimation and damaging its ability to segmentation.

\begin{table}[!t]
\centering
\resizebox{\linewidth}{!}
{
\begin{tabular}{l|l|ccc}
\toprule
Depth Extractors & Depth Supervision & DAVIS-16 & FBMS & Long. \\\hline
Monodepth2~\cite{godard2019digging} & - & 77.1  & 73.7 & 65.2  \\ 
 & Consistent Depth & \color{black}{\footnotesize$+0.4$} & \color{black}{\footnotesize$+3.2$} & \color{black}{\footnotesize$+7.9$} \\ 
 & Pseudo Depth & \color{red}{\footnotesize$-1.4$} & \color{black}{\footnotesize$+0.5$} & \color{black}{\footnotesize$+2.9$} \\  \hline \hline

LiteMono~\cite{zhang2023lite} & - & 76.8  & 79.0 & 68.1  \\ 
 & Consistent Depth & \color{black}{\footnotesize$+2.0$} & \color{black}{\footnotesize$+1.5$} & \color{black}{\footnotesize$+6.3$} \\ 
 & Pseudo Depth & \color{black}{\footnotesize$+0.7$} & \color{red}{\footnotesize$-0.1$} & \color{red}{\footnotesize$-2.6$} \\  \hline \hline

ZoeDepth~\cite{bhat2023zoedepth} & - & 79.9  & 76.4 & 64.0  \\ 
 & Consistent Depth & \color{black}{\footnotesize$+0.5$} & \color{black}{\footnotesize$+4.7$} & \color{black}{\footnotesize$+9.5$} \\ 
 & Pseudo Depth & \color{red}{\footnotesize$-0.4$} & \color{black}{\footnotesize$+0.5$} & \color{black}{\footnotesize$+1.9$} \\  \hline \hline

\bottomrule
\end{tabular}}
\caption{\textbf{TTT with depth supervision from depth predictor in DAVIS-16~\cite{perazzi2016benchmark}, FBMS~\cite{ochs2013segmentation},
Long-Videos~\cite{liang2020video} datasets.} $\mathcal{J}$ is reported for comparison. Results that 
the dropped after TTT are masked as {\color{red}{red}}. `\textit{Consistent Depth}` denotes self-supervised learning via consistent depth map prediction. `\textit{Pseudo Depth}` denotes supervised learning via pseudo depth map supervision.}
\label{tab:supp_loss}
\end{table}

\section{Further study on Naive TTT (TTT-N) strategy}
\label{sec:tttn}
We compare with other TTT methods~\cite{wang2020tent,schneider2020improving,sun2020test,gandelsman2022test} following the proposed TTT-LTV strategy~(\textit{c.f.} Section~\ref{sec:depth_aware_ttt}) and show the result in Table 4 in the main paper.
Here, we further compare with them following the naive image-based TTT strategy~(\textit{c.f.} TTT-N in Section~\ref{sec:depth_aware_ttt}). 
As shown in Table~\ref{tab:supp_ttt}, other TTT methods can not obtain consistent improvement in different datasets following the TTT-N strategy, which is the same as in the TTT-LTV strategy. Although the improvement of our method is not as obvious as that in the TTT-LTV strategy, it is more stable than others. It demonstrates that depth-aware test-time training is necessary in ZSVOS.

\begin{table*}[!t]
\centering
{
\begin{tabular}{l|l|ll|ll|ll|ll|ll}
\toprule
{Method} &  {TTT-N} & \multicolumn{2}{c|}{DAVIS-16} & \multicolumn{2}{c|}{FBMS} & \multicolumn{2}{c|}{Long.} & \multicolumn{2}{c|}{MCL} & \multicolumn{2}{c}{STV2} \\ 
 & & $\mathcal{J}$ & $\mathcal{F}$ & $\mathcal{J}$ & $\mathcal{F}$ & $\mathcal{J}$ & $\mathcal{F}$ & $\mathcal{J}$ & $\mathcal{F}$  & $\mathcal{J}$ & $\mathcal{F}$  \\ \hline
Baseline  & - & 75.9 & 77.5 & 75.1 & 76.5 & 63.9 & 67.5 & 57.3 & 70.8 & 61.5 & 70.4 \\ 
TENT~\cite{wang2020tent}  & \checkmark & \color{red}{\footnotesize$-0.2$} & \color{red}{\footnotesize$-0.2$} & \color{black}{\footnotesize$+0.2$} & \color{black}{\footnotesize$+0.3$} & \color{black}{\footnotesize$+0.4$} & \color{black}{\footnotesize$+0.3$} & \color{black}{\footnotesize$+0.7$} & \color{black}{\footnotesize$+0.3$} & \color{red}{\footnotesize$-0.1$} & \color{red}{\footnotesize$-0.1$} \\
BN~\cite{schneider2020improving}  & \checkmark & \color{red}{\footnotesize$-0.1$} & \color{red}{\footnotesize$-0.1$} & \color{black}{\footnotesize$+\textbf{0.4}$} & \color{black}{\footnotesize$+\textbf{0.6}$} & \color{black}{\footnotesize$+0.7$} & \color{black}{\footnotesize$+0.4$} & \color{black}{\footnotesize$+1.0$} & \color{black}{\footnotesize$+0.8$} & \color{black}{\footnotesize$+0.1$} & \color{black}{\footnotesize$+0.1$} \\ \hline
TTT-Rot~\cite{sun2020test}  & - & 75.3 & 76.2 & 75.4 & 77.2 & 59.1 & 62.8 & 57.7 & 70.5 & 66.4 & 73.3 \\
& \checkmark & \color{red}{\footnotesize$-0.1$} & \color{red}{\footnotesize$-0.1$} & \color{black}{\footnotesize$+0.1$} & \color{black}{\footnotesize$+0.3$} & \color{black}{\footnotesize$+0.8$} & \color{black}{\footnotesize$+1.1$} & \color{black}{\footnotesize$+1.0$} & \color{black}{\footnotesize$+0.9$} & \color{red}{\footnotesize$-0.4$} & \color{red}{\footnotesize$-0.1$} \\ \hline
TTT-MAE~\cite{gandelsman2022test} & - & 73.5 & 74.1 & 74.6 & 75.7 & 64.4 & 67.5 & 55.7 & 66.8 & 62.2 & 70.2 \\
& \checkmark &  \color{red}{\footnotesize$-0.2$} & \color{red}{\footnotesize$-0.1$} & \color{red}{\footnotesize$-0.1$} & \color{red}{\footnotesize$-0.1$} & \color{black}{\footnotesize$+0.3$} & \color{black}{\footnotesize$+0.2$} & \color{black}{\footnotesize$+0.1$} & \color{black}{\footnotesize$+0.2$} & \color{black}{\footnotesize$+0.1$} & \color{black}{\footnotesize$+0.1$} \\ \hline
Ours  & - & 77.1 & 78.4 & 73.7 & 75.8 & 65.2 & 68.0 & 53.5 & 66.2 & 61.5 & 69.2 \\ 
& \checkmark & \color{black}{\footnotesize$+\textbf{0.3}$} & \color{black}{\footnotesize$+\textbf{0.3}$} & \color{black}{\footnotesize$+0.1$} & \color{black}{\footnotesize$+0.3$} & \color{black}{\footnotesize$+\textbf{1.3}$} & \color{black}{\footnotesize$+\textbf{1.5}$} & \color{black}{\footnotesize$+\textbf{1.9}$} & \color{black}{\footnotesize$+\textbf{1.5}$} & \color{black}{\footnotesize$+\textbf{1.0}$} & \color{black}{\footnotesize$+\textbf{1.2}$} \\ \hline

\end{tabular}}
\caption{\textbf{Comparisons with state-of-the-art test-time training method on DAVIS-16~\cite{perazzi2016benchmark}, FBMS~\cite{ochs2013segmentation}, Long-Videos~\cite{liang2020video}, MCL~\cite{kim2015spatiotemporal}, and SegTrackV2~\cite{li2013video} datasets.}  Results that 
the dropped after TTT are masked as {\color{red}{red}}. The most significant improvement is marked as \textbf{bold}. 
}
\label{tab:supp_ttt}
\end{table*}

\section{Impact of the depth prediction loss weight $\lambda$}
\label{sec:lambda}
We use a hyper-parameter $\lambda$ to balance the two losses as described in Equation~\ref{eqn:loss}. We choose different $\lambda$ and find it is important in learning depth-aware features. As shown in Table~\ref{tab:supp_lambda}, a larger $\lambda$ allows the model to learn stronger 3D knowledge during the training-time training, which leads to better results when the model is directly applied to the test videos. However, the well-trained image encoder cannot benefit from the proposed self-supervised task consistently at test time. 
Finally, we choose $\lambda=0.1$ since it performs well both with and without TTT.

\begin{table}[!t]
\centering
{
\begin{tabular}{l|l|ccc}
\toprule
$\lambda$ & TTT & DAVIS-16 & FBMS & Long. \\\hline
1 & - & 77.5  & 75.4 & 65.5 \\ 
& \checkmark & \color{red}{76.7}  & \color{red}{72.4} & 67.4 \\ \hline\hline

0.1 & - & 77.1  & 73.7 & 65.2 \\ 
& \checkmark & \textbf{77.5}  & 76.9 & \textbf{73.1} \\ \hline\hline
 
0.01 & - & 76.1  & 73.8 & 64.6 \\ 
& \checkmark & 76.7  & \textbf{77.9} & 72.6 \\ 
\bottomrule
\end{tabular}}
\caption{\textbf{Impact of the depth prediction loss weight $\lambda$ in DAVIS-16~\cite{perazzi2016benchmark}, FBMS~\cite{ochs2013segmentation}, 
Long-Videos~\cite{liang2020video} datasets.} $\mathcal{J}$ is reported for comparison. Results that 
the dropped after TTT are masked as {\color{red}{red}}.
The best result is marked as \textbf{bold}. 
}
\label{tab:supp_lambda}
\end{table}

\section{Sampling strategy for densely annotated videos: DAVIS-16~\cite{perazzi2016benchmark} and SegTrackV2~\cite{li2013video}}
\label{sec:sampling}

DAVIS-16~\cite{perazzi2016benchmark} and SegTrackV2~\cite{li2013video} are densely annotated, while FBMS~\cite{ochs2013segmentation}, 
Long-Videos~\cite{liang2020video}, and MCL~\cite{kim2015spatiotemporal} are annotated once every few frames.
We perform TTT frame-by-frame on FBMS, Long-Videos, and MCL. As for DAVIS-16 and SegTrackV2, 
we first divide the video frames into 10 video clips, which means that the interval of consecutive frames in each clip is 10,
and then perform TTT by sampling a single frame from each clip. As shown in Table~\ref{tab:supp_frame_rate}, the performance may drop when performing TTT without the sampling strategy. Conducting the sampling strategy allows model training from more diverse input which helps to combat overfitting.
Similarly, Figure~\ref{fig:ttt_epoch} in the main paper shows that training too many epochs in the same frame may drop the performance in sparsely annotated video.

\begin{table}[!t]
\centering
{
\begin{tabular}{l|cc}
\toprule
Clip Nums & DAVIS-16 & STV2 \\\hline
- & 77.1  & 61.5 \\ \hline\hline
1 & \color{red}{\footnotesize$-3.1$}  & \color{black}{\footnotesize$+2.8$}  \\ 
5 & \color{red}{\footnotesize$-0.1$}  & \color{black}{\footnotesize$+3.1$} \\ 
10 & \color{black}{\footnotesize$+0.4$}  & \color{black}{\footnotesize$+4.4$} \\ 
20 & \color{black}{\footnotesize$+0.5$} & \color{black}{\footnotesize$+2.6$} \\ 
 
\bottomrule
\end{tabular}}
\caption{\textbf{Sampling strategy for densely annotated videos: DAVIS-16~\cite{perazzi2016benchmark} and SegTrackV2~\cite{li2013video} datasets.} $\mathcal{J}$ is reported for comparison. Results that 
the dropped after TTT are masked as {\color{red}{red}}.}
\label{tab:supp_frame_rate}
\end{table}

\section{Extending DATTT to existing ZSVOS method}
\label{sec:extend}
Our approach operates independently of other ZSVOS methods. For instance, we utilize HFAN~\cite{pei2022hierarchical} as the baseline model for our DATTT approach. HFAN incorporates additional feature alignment modules for both appearance and motion features. As demonstrated in Table \ref{tab:hfan}, our proposed depth-aware architecture and TTT strategy each yield noticeable enhancements.

\begin{table}[!t]
\centering
{
\begin{tabular}{l|ccc}
\toprule
  & DAVIS-16 & FBMS & Long. \\\hline
Baseline  & 77.8  & 73.6 & 64.6  \\  
+ Ours Architecture  & 78.6  & 74.1 & 65.8  \\  
+ Ours TTT Strategy  & \bf 78.8  & \bf 78.2 & \bf 71.7  \\  
\hline
\end{tabular}}
\caption{\textbf{Using HFAN~\cite{pei2022hierarchical} as the baseline model on DAVIS-16~\cite{perazzi2016benchmark}, FBMS~\cite{ochs2013segmentation}, Long-Videos~\cite{liang2020video} for TTT.} $\mathcal{J}$ is reported for comparison. 
}
\label{tab:hfan}
\end{table}

\section{Additional visual results}
\label{sec:vis}
\begin{figure*}[!th]
\centering
\includegraphics[width=\linewidth]{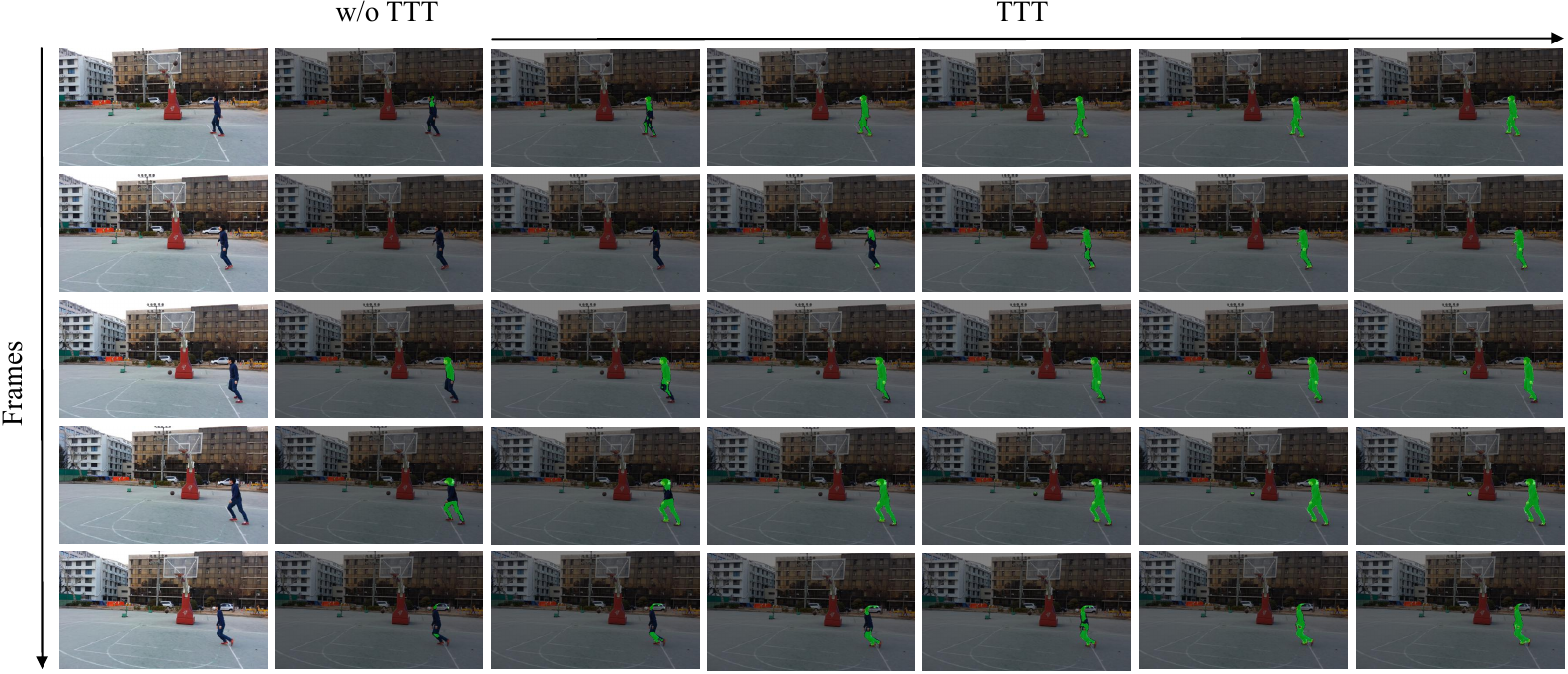}

\vspace{2em}
\includegraphics[width=\linewidth]{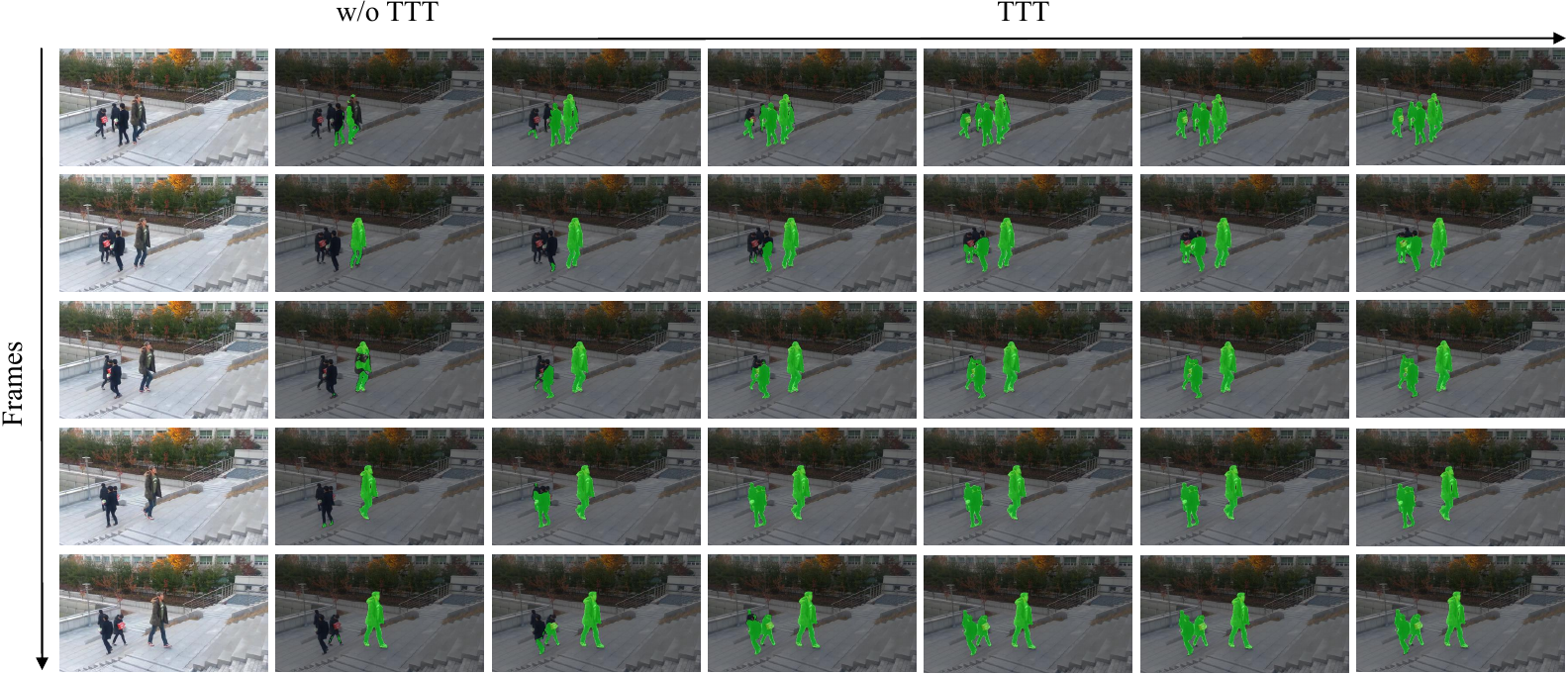}

\vspace{2em}
\includegraphics[width=\linewidth]{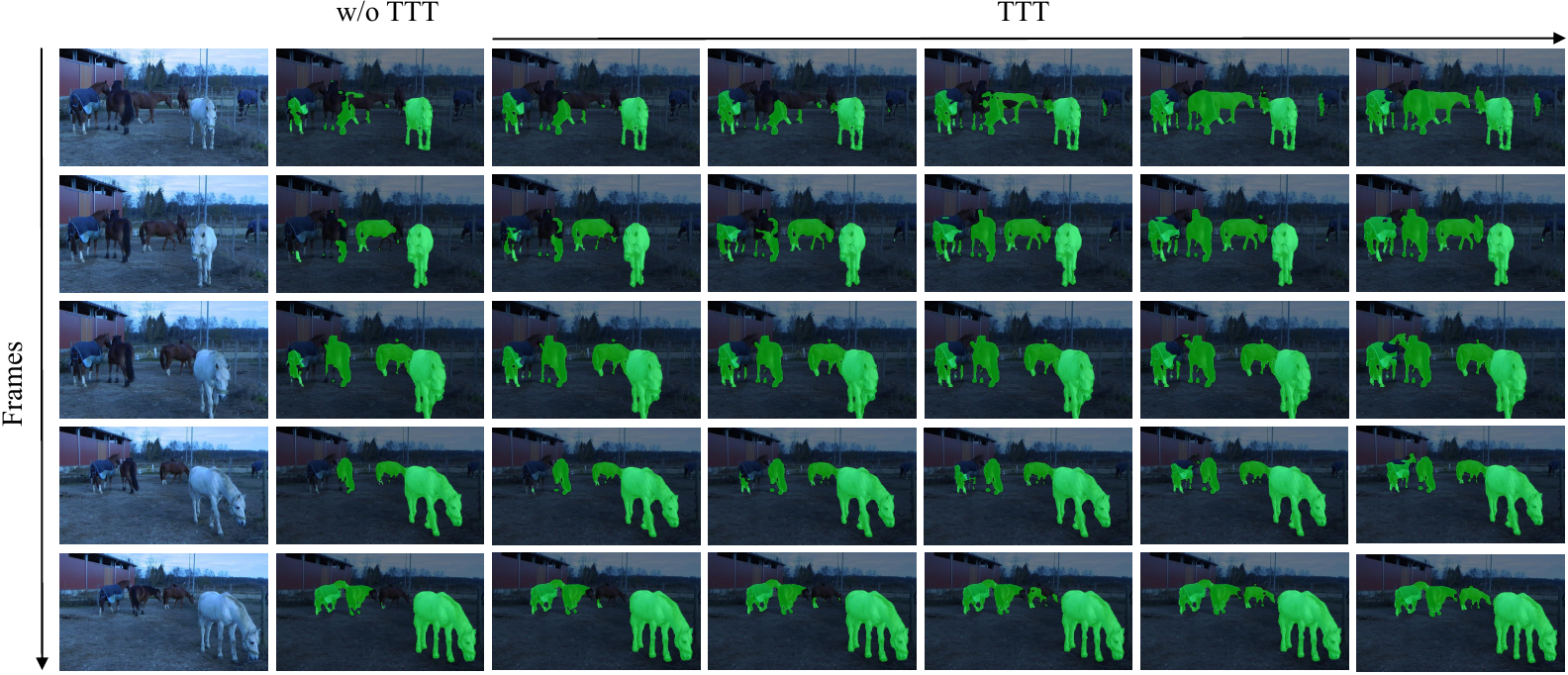}

\caption{\textbf{Additional visual results.} The background in the results is dimmed for better visualization.
}
\label{fig:supp_visualization}
\end{figure*}

We provide more visual results similar to Figure~\ref{fig:visualization}~(main paper) in Figure~\ref{fig:supp_visualization}. The pre-trained model struggles to handle these videos at first, and then clear improvements are observed after performing TTT. The proposed method works well in both single-object and multi-object scenarios.

\end{document}